# A Bagging and Boosting Based Convexly Combined Optimum Mixture Probabilistic Model

*Mian Arif Shams Adnan*
*Department of Mathematics and Statistics, Bowling Green State University,*
*Bowling Green, Ohio 43402, USA*
*maadnan@bgsu.edu*
*and*
*H. M. Miraz Mahmud*
*Dhaka, Bangladesh*

## ABSTRACT

Unlike previous studies on mixture distributions, a bagging and boosting based convexly combined mixture probabilistic model has been suggested. This model is a result of iteratively searching for obtaining the optimum probabilistic model that provides the maximum p value.

## 1. Introduction

Mixture distribution was first coined in 1894. Earlier works on this topic are due to Houghton et al. (1990), Mahlman (1997), Ahsan et al (2008), Ali (1990, 2003), Quadir et al (2002), Gupta and Kundu (2001), Mudholkar and Shrivastava (1993) and Nadarajah (2005). However, Frigessi et al. (2002), Mendes and Lopes (2004), Behrens et al. (2004) have developed some mixture models. The drawback with all the aforementioned approaches is the prior specification of a parametric model for the bulk of the distribution (and associated weight function where appropriate). Tancredi et al. (2006) has proposed a semi-parametric mixture model, A. MacDonald *et al* proposed a flexible model which includes a non-parametric smooth kernel density estimator below some threshold accompanied with the PP model for the upper tail above the threshold. A mixture of hybrid-Pareto has been carried by Carreau and Bengio (2009). Patrizia Ciarlini *et al* (2004), Maurice Cox *et al* have introduced the use of a probabilistic tool, a mixture of probability distributions, to represent the overall population in a temperature comparison. This super-population is defined by combining the local populations in given proportions. The mixture density function identifies the total data variability, and the key comparison reference value has a natural definition as the expectation value of this probability density.

Adnan *et al* (2016, 2012) demonstrates some new ways of generating the properties of a probability distribution and a new approach of the of good ness of fit test for probability models. The following paper is an initial step of finding or generating a mixture probabilistic model before the generalized version goodness of fit test for mixture distribution. In this study an attempt has been made to fit an appropriate probability model for a real-life data. Section 2 demonstrates the proposed statistical methodologies to explore a suitable probabilistic model. The real-life data and its data analysis is explained in section 3. This sector shows how to find a bagging and boosting based optimum mixture model for the said region. Final section draws the conclusion.

## 2. Methods and Methodology

The distribution function of a random variable $x$, $F(x)$, can be estimated as a mixture of single distributions. Mixtured distribution can be formed with weights $(1-p)$ and $p$ (where, p refers p-value). If we get the higher p-value for the goodness of fit test in case of the mixture extreme value distribution, and that p-value is greater than those of the other distributions, then we can say that the mixture extreme value distribution is the best probabilistic model for the observed data. So, the mixture model of the extreme value distribution with weights $(1-p)$ and $p$ is given as of the following form

$$F(x) = (1-p) * F_1(x) + p * F_{1'}(x)$$

where, $F_1(x)$ is the cumulative density function of the distribution with the estimated value of location parameter, $F_{1'}(x)$ is the cumulative density function of another distribution with changing the value of location and/or scale parameter(s). Here, $F(x)$ is a convexly combined model that gives optimum p- value after several iterations. It is also called a Bagging and Boosting based probabilistic model.

## 3. Analysis of the Data

Natural disasters like floods, tornadoes, tropical cyclones, heat and cold wave cause tremendous loss of property all over the world. Most of the natural disasters are either due to weather or are triggered due to weather related processes. Weather events are meteorological occurrences that cause weather conditions to degrade from the "ideal" weather condition. The Global Climate Risk Index 2009 showed Bangladesh as the most vulnerable to extreme weather events as a result of climate change due to its geographic location, flat and low-lying topography. Climate change, environmental degradation in Bangladesh has been occurring faster than the past two decades. As more than half of the land of Bangladesh is less than 20 feet above sea level, Bangladesh faces a double threat: one is rising sea level as a result of the melting ice caps and glaciers, and the other is extreme climatic events, like cyclones and heavy rain. Climate change is changing most of our traditional agricultural practices as the seasonal cycle and rainfall pattern have changed, droughts have become more frequent, violent stresses of cyclones, earthquakes, prolonged floods, salt water intrusion are increasing day by day. The average temperature has increased in the summer while winter season has shortened.

Some projections about climate change are of great concern for Bangladesh such as: temperature rise would be 1.30 C by 2030 and 2.60 C by 2070, the sea level rise would be up to three feet and a greater part of the costal area would be inundated. As a consequence, 17 percent of total cultivable land would be affected. Bangladesh has in the last few years experienced cyclones, floods and heat waves that have killed thousands and left millions more homeless. Sea Surface Temperature (SST) fulfils one of the major preconditions for the formation of depressions and low-pressure systems in the Bay of Bengal. The increased frequency of rough sea conditions caused by depressions is actually a consequence of climatic change. The fourth assessment report of the Inter-Governmental Panel on Climate Change also supports this argument.

Maximum temperature, minimum temperature and the number of hot days is important determinants of weather condition of a region. An appropriate probabilistic model may serve the purpose.

In Jessore Station fifty-one years of maximum temperature data during 1948 to 1998 have been used to fit probability models. The missing value of the data has been estimated on the basis of average temperature of two adjacent years. The minimum, maximum, location and dispersion of the yearly maximum daily temperatures are 36.1° C, 43.8° C, 40.058° C and 2.1892° C respectively. A time series of the maximum temperatures is portrayed below.

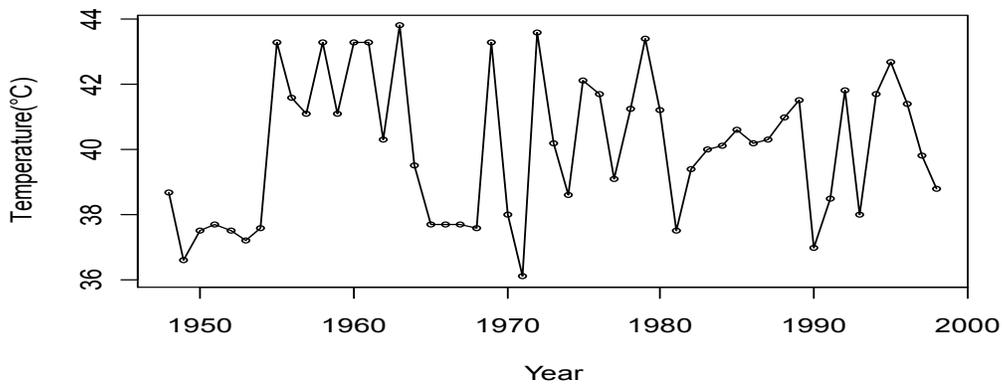

Figure 1 : Yearly maximum temperature recorded at Jessore station

The generalized extreme value distribution of Jenkinson (1955) is widely being used for modeling extremes of natural phenomena, and it is of considerable importance in hydrology. The magnitude of extreme events may be taken, to a reasonable approximation, as being distributed according to one of the three extreme-value distributions defined by Fisher and Tippett (1928). The three extremal models are Gumbel distribution (extreme value type I (EV1) distribution), Frechet distribution (extreme value type II (EV2) distribution) and the Weibull distribution (extreme value type III (EV3) distribution). Suppose $x_1, x_2, ..., x_n$ denote annual maximum daily temperature for $n$ years from a given location. The method of maximum likelihood can be used to fit extreme value distribution to these data. This maximization can be performed using a quasi-Newton iterative algorithm. The standard errors of the estimates can also be computed by inverting the Fisher information matrix. To obtain an accurate probability distribution of extreme temperature at the following information the R programming has been used. To search the appropriate probability distribution for extreme temperature data, the chi-square goodness of fit test, Kolmogorov-Smirnov test, Q-Q plot, P-P plot, Density plot have been used. In case of the class frequency less than five the class frequency to that class has been amalgamated with the frequency in the immediately upper class. The maximum likelihood estimates of the parameters of Generalized Extreme Value Distribution of the extreme temperatures for the aforementioned region are 39.22° C as location, 2.182° C as scale and -0.237 as shape. The same estimates for the parameters for the Gumbel, Weibull, Frechet distributions are also found.

The Q-Q plot on the basis of annual maximum daily temperature data for all of the four distributions have been found. The four Q-Q plots cannot conclude the origin of the data since for all cases the points fall approximately along with $45^0$ reference line.

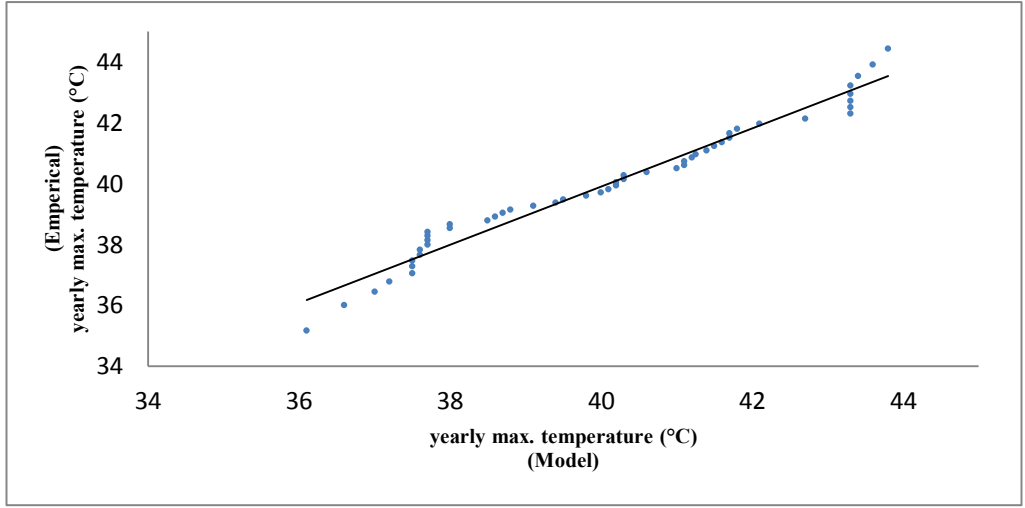

Figure 2: Q-Q plot for the Generalized Extreme Value distribution.

Letting the null and alternative hypothesis to be the data following generalized extreme value distribution the Kolmogorov-Smirnov test statistic is calculated as $\widehat{D_n} = \sup_x |F_n(x) - F(x)| = 0.102$ whereas the critical value is $D_{n,\alpha} = D_{51, 0.05} = \frac{0.89}{\sqrt{n}} = 0.1246$. Since, $\widehat{D_n} < D_{n,\alpha}$, we do not reject the null hypothesis at 1% level of significance. Therefore, the annual maximum daily temperature data follows generalized extreme value distribution. Similarly, it is noticed that the data does not follow Gumbel, Weibull or even Frechet distribution at the same level of significance along with the values of the same test statistic as $0.16, 0.1246$ and $0.15$ respectively.

Similarly, p- value of chi-square statistic on the basis of 51 years of maximum temperature data is 0.05416. So, the null hypothesis is not rejected at 5% level of significance. Therefore, annual maximum daily temperature data follows Extreme Value distribution. It is also evident that the data does not come from Gumbel, Weibull or Frechet with the lower p-values 0.00575, 0.003818 and 0.01458 respectively.

The P-P plot of the sorted values (in ascending order) of the observed annual maximum daily temperature versus expected quantiles $y_i$ determined by $y_i = \left(\frac{i-0.5}{n+1}\right)$ plotted. Figure 4 represents (for the extreme value distribution) the P-P plot on the basis of annual maximum daily temperature data where the plot shows that the points fall approximately along with $45^0$ reference line which means that these data follows extreme value distribution.

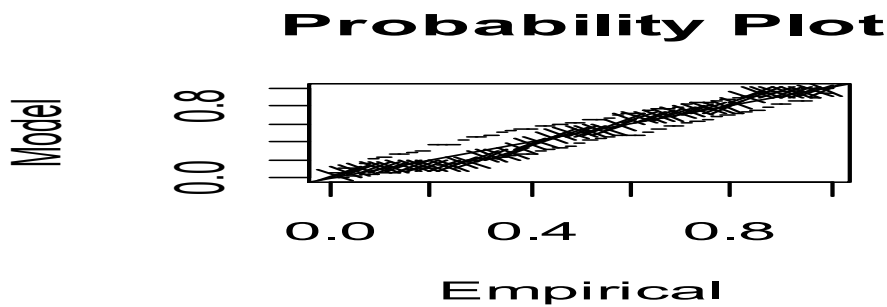

Figure 3: Probability plot for Extreme Value distribution.

In the following figure, density plot also shows that data plots are approximately close to the original density which is an assurance that the data follows extreme value distribution.

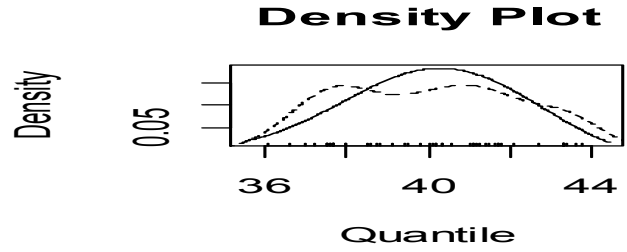

Figure 4: Density plot for Extreme Value distribution.

Since bimodality of the distribution of extreme temperatures indicates the possibility of the extreme of a mixture of two extreme value distributions, we have tried to fit an appropriate mixture of two extreme value distributions in this paper. As such, the cumulative density function of mixtured extreme value distribution is given by

$$F(x) = (1 - 0.05416) * F_1(x) + 0.05416 * F_{1'}(x).$$

Now, letting the null hypothesis to be the data following the mixture extreme value distribution against the alternative hypothesis not to be true, the value of chi-square test statistic on the basis of 51 years of maximum temperature data under the postulated mixture model for different values of location parameter are shown in the Table 1 of the appendix. Table 1 shows that p-value for the mixtured extreme value distribution is 0.0549001 along with the location parameter is 39.50° C. Moreover, p-value increases as the value of location parameter increases up to the point i.e., 43.76° C, then it increases slowly around 0.08301 for values 43.77° C to 43.78° C and after that p-value decreases. That is, p-value is maximum ($\equiv$ 0.0830111) for the value of location parameter 43.78° C. So, cumulative density function of the optimum mixtured extreme value model with the location parameters 39.22° C and 43.78° C is given by

$$F(x) = (1 - 0.05416) * exp\left[-\left\{1 - 0.237\left(\frac{x-39.22}{2.182}\right)\right\}^{\frac{1}{0.237}}\right] + 0.05416 *$$

$$exp\left[-\left\{1 - 0.237\left(\frac{x-43.78}{2.182}\right)\right\}^{\frac{1}{0.237}}\right].$$

$$\therefore \frac{d}{dx}F(x) = \frac{d}{dx}\left[(1 - 0.05416) * exp\left[-\left\{1 - 0.237\left(\frac{x-39.22}{2.182}\right)\right\}^{\frac{1}{0.237}}\right]\right.$$

$$\left. + 0.05416 * exp\left[-\left\{1 - 0.237\left(\frac{x-43.78}{2.182}\right)\right\}^{\frac{1}{0.237}}\right]\right]$$

gives maximum p-value (=0.0830111) for chi-square goodness of fit test. So, the null hypothesis of mixture model with two locations 39.22° C and 43.78° C is more strengthenly accepted with relatively higher p-value.

However, there might have some inquiry that another mixture model like mixture model based on four afore mentioned distributions (extreme value, weibull, gumbel, frechet) might predict well. But truly, it will not be a suitable one, since, except extreme value distribution, no distribution among the rest of the three distributions does not fit to the observed data at all.

Now, for fixed locations $\mu_1 = 39.22$, $\mu_2 = 43.78$, and shapes $\xi_1 = -0.237, \xi_2 = -0.237$; if $\sigma_2$ is increased then p-value decreases (Table-2 of appendix), and if $\sigma_2$ is decreased then p-value increases (Table-3).

Again for fixed locations $\mu_1 = 39.22$, $\mu_2 = 43.78$, shapes $\xi_1 = -0.237$, $\xi_2 = -0.237$; if $\sigma_1$ is decreased then p-value also decreases (Table-4), and if $\sigma_1$ is increased then p-value also increases but after certain value of $\sigma_1$ ( = 3.15) p-value decreases, i.e., we get the maximum p-value=0.18169 at $\sigma_1$=3.15 (Table-5).

Finally, for those fixed locations if $\sigma_2$ increases then p-value decreases (Table-6); If $\sigma_2$ decreases then p-value increases but after certain value of $\sigma_2$ (=.4) p-value decreases, i.e., we get the maximum p-value=0.2278 at $\sigma_2$ =.4 (Table-7). So finally, $\mu_1 = 39.22$, , $\mu_2 = 43.78$, $\sigma_1$=3.15, $\sigma_2$ =.4 gives the maximum p-value.

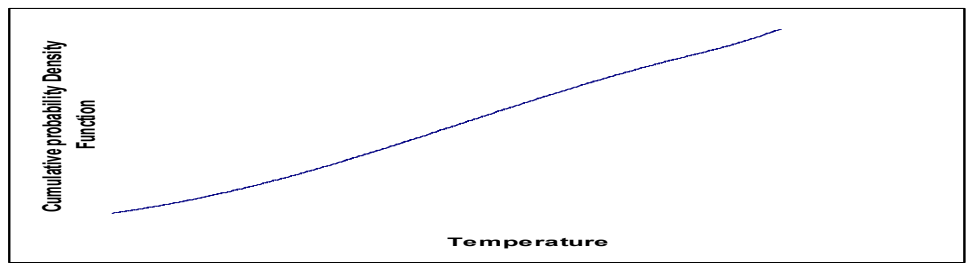

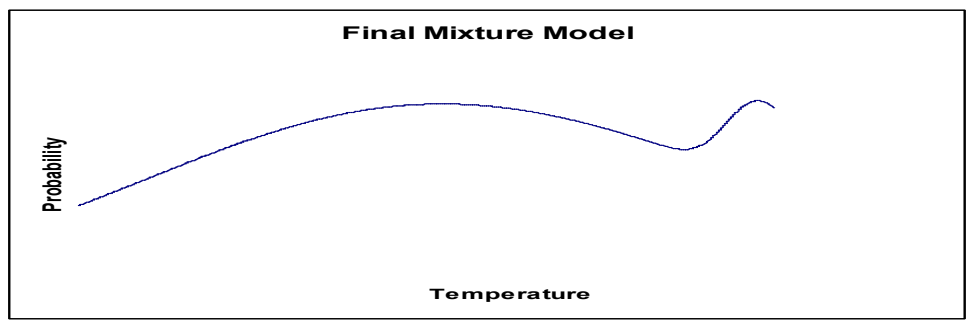

Therefore, the final mixture model for temperatures of the aforementioned region is as follows:

$$F(x) = (1 - 0.05416) * exp\left[-\left\{1 - 0.237\left(\frac{x-39.22}{3.15}\right)\right\}^{\frac{1}{0.237}}\right] + 0.05416 * exp\left[-\left\{1 - 0.237\left(\frac{x-43.78}{.4}\right)\right\}^{\frac{1}{0.237}}\right]; 36.1 \leq x \leq 43.8.$$

## Conclusion

Unlike previous studies on mixture distributions, a bagging and boosting based convexly combined mixture probabilistic model is found as a result of iteratively searching for obtaining the optimum probabilistic model that provides the maximum p value. For extreme temperatures, a very simple mixture probabilistic model approach has been applied in case of Jessor Division of Bangladesh. The optimum mixture of extreme value distributions has been found to be relatively more appropriate in case of the said region of Bangladesh. The authors are trying to develop several bagging and boosting based probabilistic and regression models for high dimensional populations.

**APPENDIX**

Table 1: Calculation of chi-square statistic and p-value of a mixtured extreme value distribution for different location parameter.

| Location ($\mu$) | Chi-square Statistic | p-value of goodness of fit test for mixture extreme value distribution |
|---|---|---|
| 39.50 | 12.3349601 | 0.0549001 |
| 40.00 | 12.2330958 | 0.0569662 |
| 43.75 | 11.1787136 | 0.0830076 |
| 43.76 | 11.1786461 | 0.0830096 |
| 43.77 | 11.1786062 | 0.0830108 |
| **43.78** | **11.1785939** | **0.0830111** |
| 43.79 | 11.1786093 | 0.0830107 |
| 43.80 | 11.1786524 | 0.0830094 |
| 43.81 | 11.1787231 | 0.0830073 |

| Table-2 | | |
|---|---|---|
| $\sigma_1$ | $\sigma_2$ | P-value |
| 2.182 | 2.182 | 0.08301 |
| 2.182 | 2.2 | .08269 |
| 2.182 | 2.3 | .08105 |
| 2.182 | 2.6 | .07699 |
| 2.182 | 2.9 | .07396 |
| 2.182 | 3 | .07312 |
| 2.182 | 3.5 | .06988 |
| Table-3 | | |
| 2.182 | 2.1 | .08450 |
| 2.182 | 2 | .08650 |
| 2.182 | 1.9 | .08872 |
| 2.182 | 1.5 | .10045 |
| 2.182 | 1 | .12809 |
| 2.182 | 0.5 | .18463 |
| 2.182 | 0.1 | .2514 |
| 2.182 | 0.05 | .25267 |

| Table-4 | | |
|---|---|---|
| 2.182 | 2.182 | 0.08301 |
| 2.1 | 2.182 | 0.06272 |
| 2 | 2.182 | 0.03899 |
| 1.9 | 2.182 | 0.01955 |
| 1.7 | 2.182 | 0.00153 |
| 1.5 | 2.182 | 0.000003 |
| Table-5 | | |
| 2.2 | 2.182 | 0.08738 |
| 2.3 | 2.182 | 0.11035 |
| 2.5 | 2.182 | 0.14622 |
| 2.6 | 2.182 | |
| 2.7 | 2.182 | 0.16783 |
| 2.8 | 2.182 | |
| 2.9 | 2.182 | 0.17837 |
| 3.00 | 2.182 | 0.18071 |
| 3.05 | 2.182 | 0.18133 |
| 3.10 | 2.182 | 0.18164 |
| 3.15 | 2.182 | 0.18169 |
| 3.2 | 2.182 | 0.18150 |
| 3.3 | 2.182 | 0.18055 |
| 3.4 | 2.182 | 0.17900 |
| Table-6 | | |
| 3.15 | 2.182 | .18169 |
| 3.15 | 2.2 | .1816 |
| 3.15 | 2.3 | .1812 |
| 3.15 | 2.5 | .1807 |
| 3.15 | 2.8 | .1804 |
| 3.15 | 3 | .1805 |
| 3.15 | 3.2 | .1806 |
| Table-7 | | |
| 3.15 | 2.182 | .18169 |
| 3.15 | 2 | .1825 |
| 3.15 | 1.8 | .1838 |
| 3.15 | 1.5 | .1868 |
| 3.15 | 1 | .1998 |
| 3.15 | .8 | .2100 |
| 3.15 | .7 | .2159 |
| 3.15 | .6 | .2215 |
| 3.15 | .5 | .2258 |
| **3.15** | **.4** | **.2278** |
| 3.15 | .3 | .2266 |
| 3.15 | .2 | .2212 |
| 3.15 | .1 | .2104 |